\pdfoutput=1
\documentclass[11pt]{article}

\usepackage[preprint]{acl}

\usepackage{times}
\usepackage{latexsym}

\usepackage[T1]{fontenc}

\usepackage[utf8]{inputenc}

\usepackage{microtype}

\usepackage{inconsolata}
\usepackage{booktabs}
\usepackage{multirow}
\usepackage{rotating}
\usepackage{enumitem}
\usepackage{tablefootnote}
\usepackage{pifont}

\usepackage{pdfpages}
\usepackage{longtable}
\usepackage{setspace}
\usepackage{etoolbox,xspace}
\usepackage{comment}

\usepackage{svg}
\usepackage{caption}
\usepackage{enumitem}
\usepackage{graphicx}
\usepackage{amssymb}
\usepackage{xspace}
\usepackage{makecell}
\usepackage{colortbl}
\usepackage{inconsolata}
\usepackage[most]{tcolorbox}
\usepackage{graphicx}

 \usepackage{pifont}
 \usepackage{todonotes}

\title{\textbf{\textsc{CuLEmo}}: Cultural Lenses on Emotion - Benchmarking LLMs  \\ for Cross-Cultural Emotion Understanding}

\author{
  \textbf{Tadesse Destaw Belay\textsuperscript{1,3}},
  \textbf{Ahmed Haj Ahmed\textsuperscript{2}},
  \textbf{Alvin Grissom II\textsuperscript{2}},
  \textbf{Iqra Ameer\textsuperscript{4}},\\
  \textbf{Grigori Sidorov\textsuperscript{1}},
  \textbf{Olga Kolesnikova\textsuperscript{1}},
  \textbf{Seid Muhie Yimam\textsuperscript{5}}
  \\
  \\
  \textsuperscript{1}Instituto Politécnico Nacional,
  \textsuperscript{2}Haverford College,
  \textsuperscript{3}Wollo University,\\
  \textsuperscript{4}Pennsylvania State University,
  \textsuperscript{5}University of Hamburg
  \\
  \small{
    \textbf{Correspondence:} \href{mailto:tadesseit@gmail.com}{tadesseit@gmail.com}
  }
}

\begin{document}
\maketitle
\begin{abstract}
NLP research has increasingly focused on subjective tasks such as emotion analysis. However, existing emotion benchmarks suffer from two major shortcomings: (1) they largely rely on keyword-based emotion recognition, overlooking crucial cultural dimensions required for deeper emotion understanding, and (2) many are created by translating English-annotated data into other languages, leading to potentially unreliable evaluation. To address these issues, we introduce Cultural Lenses on Emotion (\textbf{CuLEmo}), the first benchmark designed to evaluate culture-aware emotion prediction across six languages: Amharic, Arabic, English, German, Hindi, and Spanish. CuLEmo comprises 400 crafted questions per language, each requiring nuanced cultural reasoning and understanding. We use this benchmark to evaluate several state-of-the-art LLMs on culture-aware emotion prediction and sentiment analysis tasks. Our findings reveal that (1) emotion conceptualizations vary significantly across languages and cultures, (2) LLMs performance likewise varies by language and cultural context, and (3) prompting in English with explicit country context often outperforms in-language prompts for culture-aware emotion and sentiment understanding. 
The dataset\footnote{\url{https://huggingface.co/llm-for-emotion}} and evaluation code\footnote{\url{https://github.com/llm-for-emotion/culemo}} is available.



\end{abstract}

\section{Introduction}
\label{sec:intro}
Despite progress in bridging language barriers \cite{ahuja-etal-2023}, large language models (LLMs) still struggle to capture cultural nuances and adapt to specific cultural contexts \cite{understanding-2024}. Ideally, multilingual LLMs can not only facilitate cross-lingual communication but also incorporate an awareness of cultural sensitivities (i.e., what is deemed acceptable, normal, or inappropriate in a given culture), integrating such knowledge to foster deeper global connections \cite{multilingual-2024}.

\begin{figure}[t]
    \centering
    \includegraphics[width=\linewidth]{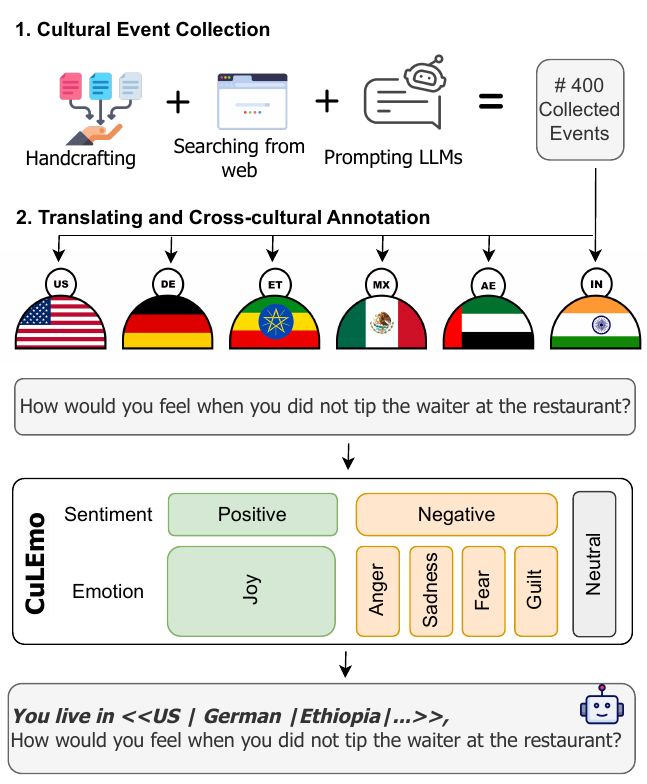}
    \caption{CuLEmo dataset creation pipeline and evaluations of LLMs in emotion and sentiment tasks.}
    \label{fig:pipline}
\end{figure}

LLMs and agent systems are employed to interact extensively with humans across applications such as customer service, healthcare, and education~\cite{Wang-2024}. To facilitate effective interaction, incorporating aspects of cognitive and emotional-social intelligence, the ability to recognize and interpret human emotions~\cite{mathur2024advancing}, can facilitate better interactions with more people. Factors such as age, cultural background, and personal experiences influence how individuals perceive and process information, particularly within subjective NLP tasks. Among these, emotion recognition (ER) and sentiment analysis (SA) are particularly sensitive to language- and culture-specific nuances~\cite{plaza-2024-emotion}.

Natural language frequently encodes emotional information~\cite{Jamin-recent-2024}. For example, consider tipping customs in restaurants: in some cultures (e.g., North America), tipping is widely practiced, whereas in China, it is rare, and in Japan, it may even be considered offensive~\cite{givi2017sentimental}. Such differences underscore the importance of culturally-aware language technologies.


Although prior work has attempted cross-lingual emotion evaluation by translating emotion-annotated data in English into other languages~\cite{tahir2023effect,de-bruyne-2023-paradox}, relying solely on translations from English can introduce incomplete or misleading insights. A more fair comparison requires the same underlying scenarios, each annotated natively across different languages and cultures. While emotion is language- and culture-dependent~\cite{plaza-2024-emotion}, comprehensive cross-cultural evaluations remain largely unexplored. 

To bridge this gap, we propose Cultural Emotion~(\textbf{CuLEmo}), a novel dataset that captures events and annotates them across multiple cultures and languages from scratch. CuLEmo enables the evaluation of multilingual LLM performance in analogous scenarios across different cultural contexts.  Figure~\ref{fig:pipline} shows the CuLEmo dataset creation and evaluation pipeline. 


Culture can manifest in 1) the language of the data itself and 2) the annotation labels (i.e., multi-culturally informed annotations) \cite{liu2024culture}. CuLEmo satisfies both conditions: it is multilingual and includes culturally grounded annotations. Indeed, the same event may evoke distinct emotional reactions in different cultures.  In light of this, we pose the following research questions (RQs): 
\begin{itemize}[noitemsep]
    \item \textbf{RQ1.} Do LLMs provide culturally aware-emotional responses?
    \item \textbf{RQ2.} Which cultures are more effectively represented in LLMs?
    \item \textbf{RQ3.} Can LLMs identify a country’s culture based on the text describing an event in the prompt?
    \item \textbf{RQ3.} Does the language of the prompt affect the ability of LLMs for culture-aware emotion understanding?
\end{itemize}

To that end, this paper makes three key contributions. First, we introduce CuLEmo, a high-quality, multicultural, and multilingual benchmark dataset. Second, we leverage CuLEmo to investigate whether widely used multilingual LLMs can capture variations in emotional expression across cultures and languages in emotion and sentiment tasks. Finally, we highlight the variation in performance on culture-aware emotion understanding when LLMs are prompted in different languages.

\section{Related Work}
\label{sec:related}
We now review related work in culturally-aware NLP, culture-oriented benchmarks, and cross-lingual study of linguistic emotional expression.
Although culture is a complex concept, most definitions of culture encompass people, groups of people, and interactions between individuals and groups \cite{liu-culture-survey-2024}. Understanding culture is important for the safety and fairness of LLMs.
\subsection{Culture-oriented Benchmarks}
Given the significance of culture in language model evaluation \cite{culturellms-2024}, researchers have proposed various culture-oriented benchmarks to explore its effects on language understanding and generation. These efforts typically involve collecting and annotating multilingual and multicultural corpora to study cultural-driven phenomena in downstream NLP tasks. For instance, prior work has examined cross-cultural user statements \cite{liu-etal-2021-visually,benchmarkingvlm-2024}, detected cultural differences and user attributes \cite{sweed-shahaf-2021-catchphrase,qian2021pchatbot}, studied multilingual moral understanding \cite{guan-etal-2022-corpus,2022-artelingo}, and addressed culture-specific time expression grounding \cite{shwartz-2022-good,2023-normsage}.

\subsection{Emotion Across Languages}
Several studies have examined multilingual emotion datasets. Some findings suggest that emotion categories can be preserved through machine translation, for example, by exploring how English-annotated emotion data translate into Finnish, French, German, Hindi, and Italian~\cite{kajava2020emotion,tahir2023effect,bianchi-etal-2022-xlm}. These works suggest that the changes in emotion labels often stem from the inherent difficulty of annotation rather than from linguistic differences. 

Conversely, other works emphasize that emotions may not be consistently preserved across different languages. \citet{language-2022} showed that typologically dissimilar languages pose challenges for cross-lingual learning with mBERT-based models. \citet{de-bruyne-2023-paradox} argued that translation could fail to capture language-specific verbalizations and connotations---especially if certain emotion keywords do not exist in a given language (e.g., there is no direct word for ``sadness'' in Tahitian, and Amharic does not have an exact term for ``surprise''). \citet{qian-etal-2023-evaluation} found that roughly half of the machine-translated outputs from English to Chinese fail to adequately preserve the original emotion, attributing these discrepancies to emotion-specific words and complex linguistic phenomena.

\subsection{Emotions and Cultures}
Recent work examines how cultural contexts shape emotional expression across languages. \citet{havaldar-etal-2023} analyze embeddings of 271 emotion keywords in English, Spanish, Chinese, and Japanese by projecting into a Valence-Arousal plane using XLM-RoBERTa \cite{xlm-roberta-2020} embeddings, finding that multilingual models embed non-English emotion words differently. \citet{multilingual-2023} analyzes Pride/Shame as a known cultural difference by prompting GPT-3.5 and GPT-4 to explore how these models handle pride and shame in the USA vs.\ Japan. They find that GPT-3.5 displays limited knowledge of culturally specific norms.  \citet{ahmad-etal-2024-generative} expands the 19 cultural questions of the work \cite{multilingual-2023} to 37 questions and evaluates ChatGPT for the low-resource Hausa language for sentiment analysis, but these evaluations are limited to 19/37 culturally relevant questions (not diverse and representative data), limited classes (Pride/Shame or positive/negative/mixed), and limited cultures (USA vs.\ Japan or Hausa).

\section{CuLEmo Dataset Preparation}
\label{sec:data_prep}
We now describe the precise steps in creating the CuLEmo dataset.

\subsection{Collecting Cultural Events}  
We manually craft scenarios, search on the web, and prompt LLMs to gather traditions, events, norms, and actions that elicit culturally different emotions across six target countries (UAE, USA, Germany, Ethiopia, India, and Mexico). 
We draw inspiration from the work of \citet{multilingual-2023} to enhance topic diversity. Language representatives are asked to propose events distinct from those of other countries in the form of emotion-oriented questions. Importantly, these events do not contain explicit emotional keywords (as typically seen in traditional emotion datasets).  We also refer to the International Survey on Emotion Antecedents and Reactions (ISEAR) \citep{scherer1994evidence} data format, \textit{"When I ... situations that cause a specific emotion"}, a well-known English dataset for emotion analysis consisting of self-reported events from around 3,000 respondents across 37 countries and five continents. Table \ref{tab:catagory} displays the ten broad categories of the CuLEmo dataset and the number of questions in each category.

\begin{table}[h]
    \centering
    \begin{tabular}{lc}
    \toprule
        \textbf{Categories} & \textbf{\# Qn.}\\
        \hline
        Family relationships & 45\\
        Social etiquette and interactions& 65\\
        Personal appearance and dress code&32 \\
        Cultural and religious practices &62 \\
        Sexual and intimate relationships &38 \\
        Professional contexts &28 \\
        Food and dining etiquette & 35\\
        Personal privacy & 25\\
        Emotional and psychological situations & 40\\
        Public behavior and norms & 30\\
        \hline
       \textbf{Total} & \textbf{400}\\
        \bottomrule
     \end{tabular}
    \caption{The ten broad categories and the number of events (\textbf{\#  Qn.} - number of questions) in the CuLEmo dataset.}
    \label{tab:catagory}
\end{table}

\subsection{Human-Adapted Translation} 
After collecting the events in English, we translate them into five target languages—Arabic, Amharic, German, Hindi, and Spanish—using Google Translate, followed by native-speaker approvals. Because the questions are simple \textit{“How do you feel when … ?”} questions and lack explicit emotion keywords, translation quality did not affect their cultural content. While we acknowledge that existing works often depend on translating emotion-annotated data from English into other languages with their labels \cite{kajava2020emotion,tahir2023effect,bianchi-etal-2022-xlm}, our translation process is done before any annotation. To ensure correctness, native speakers evaluate the translations. While most translations were acceptable, a few adjustments were made, e.g., to fix gender references and timing expressions for Amharic.

\subsection{Language and Cultures Covered}
Several factors guide our choice of languages, ensuring a broad range of cultural norms and concepts: (1) typological variety (five languages with four scripts), (2) geographical diversity (eastern vs.\ western contexts), (3) resource availability (low- vs.\ high-resource languages), and (4) the availability of native speakers for translation reviews.


\subsection{CuLEmo Annotation}\label{sec:anno}

We use Amazon Mechanical Turk (MTurk) for most of our annotations, ensuring at least five native-speaker annotators per instance from the respective targeted country. We use a customized POTATO \cite{pei-etal-2022-potato} annotation tool for languages lacking sufficient MTurk annotators (e.g., Amharic) and recruit local native speakers who met our criteria. Annotators are fairly compensated \$12/hr (better than the Prolific\footnote{\url{https://www.prolific.com/}}'s minimum annotation wage, which is \$9/hr). Where no majority vote emerges among five annotators, we assign two additional annotations and use that majority vote. 

\begin{figure}[h]
    \centering
    \includegraphics[width=\linewidth]{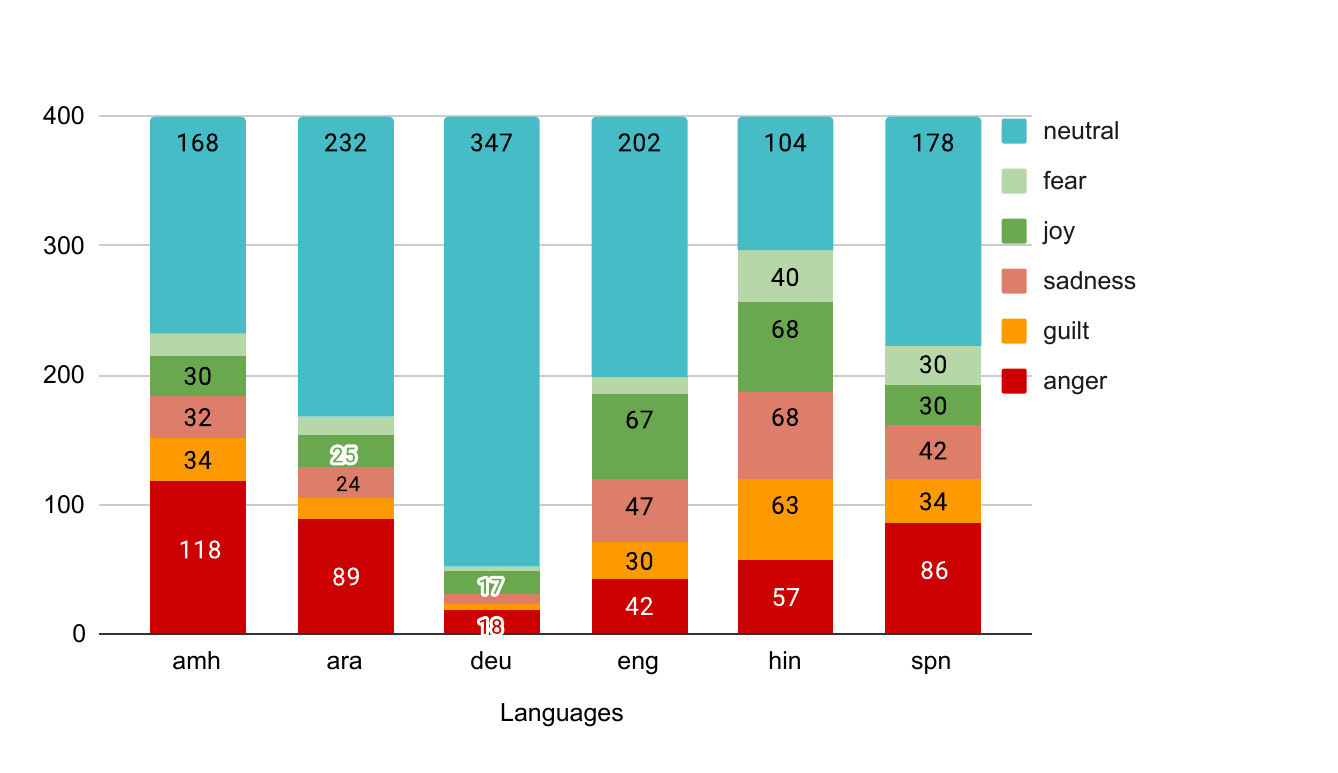}
    \caption{Emotion label distribution across countries/languages: the number of instances in each emotion label across languages from a total of 400 events.} 
    \label{fig:distr}
\end{figure}

Figure~\ref{fig:distr} illustrates the distribution of emotion labels. We use six categories: \textit{joy}, \textit{fear}, \textit{sadness}, \textit{anger}, \textit{guilt}, and \textit{neutral} (no specific emotion). These labels were adapted from \citet{de2019towards}, where five categories were clustered, plus a neutral class. Our annotation guidelines group related labels as “helper” categories—for instance, “love” and “happy” under \textit{joy}, and “shame” under \textit{guilt}—to assist annotators in selecting the most appropriate coarse-grained label. Examples from the dataset are provided in Appendix \ref{app:example}.

\begin{figure}[!h]
    \centering
    \includegraphics[width=0.95\linewidth]{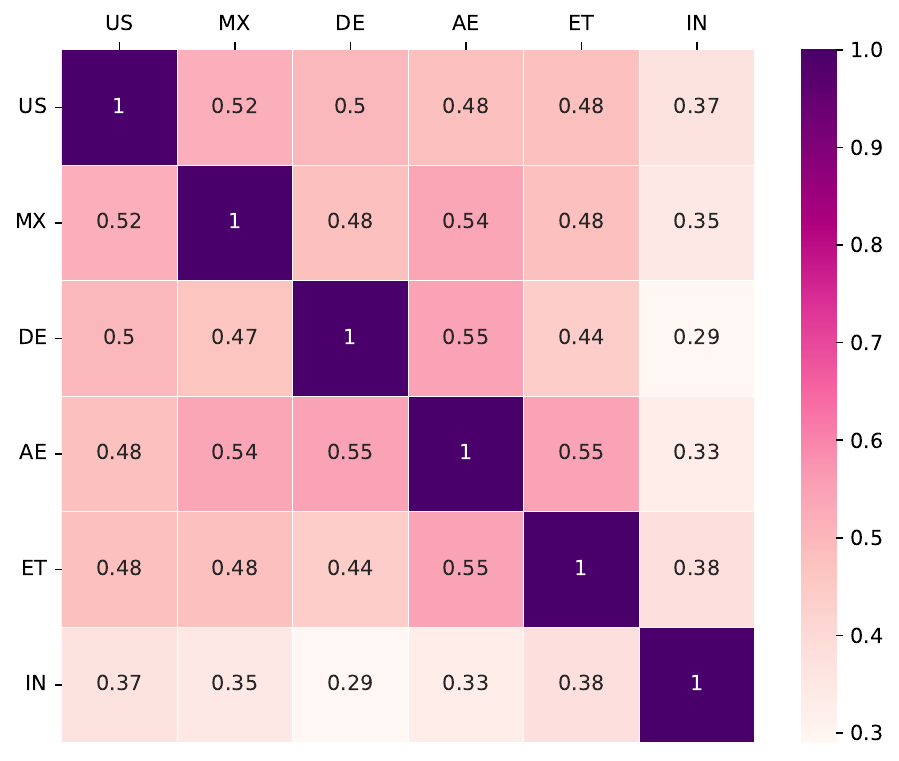}
    \caption{Pairwise emotion label agreements across countries/languages (ordered by their average agreement). Abbreviations: US = USA, MX = Mexico, DE = Germany, AE = UAE, ET = Ethiopia, and IN = India.}  
    \label{fig:cross}
\end{figure}

\noindent\textbf{Pairwise Label Agreements Across Countries}  We further examine label differences across countries by computing pairwise agreements after majority voting (Figure~\ref{fig:cross}). Ethiopia and the United Arab Emirates exhibit the highest agreement at 55\%, along with Germany and the United Arab Emirates; both exhibit a high \textit{neutral} class (Figure \ref{fig:distr}), while Germany and India show the lowest agreement, at 29.0\%. Labels from India thus diverge from those of other countries.

\section{Experimental Setup}
\label{sec:experiments}
\subsection{Task Formulation}\label{task}
To investigate the emotional comprehension capabilities of LLMs, we examine how they associate different cultures of a country with their corresponding languages. Specifically, we explore culture-aware emotion understanding via two main tasks: (1) emotion prediction and (2) sentiment analysis. All tested models are instruction-fine-tuned, except for the Aya-expanse model. We also experiment with prompts that do and do not include explicit country context, using the phrase \texttt{"You live in <<country name>>,"} (where \texttt{<<country name>>} is one of the six targeted countries: UAE, USA, Ethiopia, Germany, India, and Mexico). Each task is framed as a text-generation problem, and the models are evaluated in a zero-shot setting.

\subsection{Model Selection}
We evaluate a variety of recent LLMs known for strong performance on standard benchmarks. We aim to include both smaller and medium-sized models, as well as open-source and proprietary models:
\begin{enumerate}[nosep]
    \item \textbf{Open Source}: LLaMA-3 (3.2-3B, 3.1-8B) \cite{llama3.1,llama3.2}, Gemma (2B, 9B) \cite{gemmateam2024gemma}, Aya (expanse-8b, 101-13B) \cite{aya-2024,dang2024ayaexpanse}, Ministral (3B, 8B) \cite{ministral}
    \item \textbf{Proprietary}: GPT (3.5, 4) \cite{gpt4technicalreport}, Gemini-1.5 \cite{gemini1.5}, Claude (3.5-sonnet, 3-opus) \cite{claude3}
\end{enumerate}

\subsection{Multilingual Prompt Construction}
To examine the impact of prompt language on model performance and to assess each model’s cultural awareness, we design both \textbf{English} and \textbf{in-language} prompts. The instruction, input text, and expected answer are in English when using English prompts. For in-language prompts, all elements are in one of the five target languages—Arabic (AR), Amharic (AM), German (DE), Hindi (HI), or Spanish (ES). Complete examples of our multilingual prompts for both emotion prediction and sentiment analysis are provided in Appendix~\ref{app:prompts}. We extract each model’s answer from its generated text using the PEDANTS tool \cite{li-etal-2024-pedants}.

\section{Result and Analysis}

\begin{figure*}[]
    \centering
    \includegraphics[width=\linewidth]{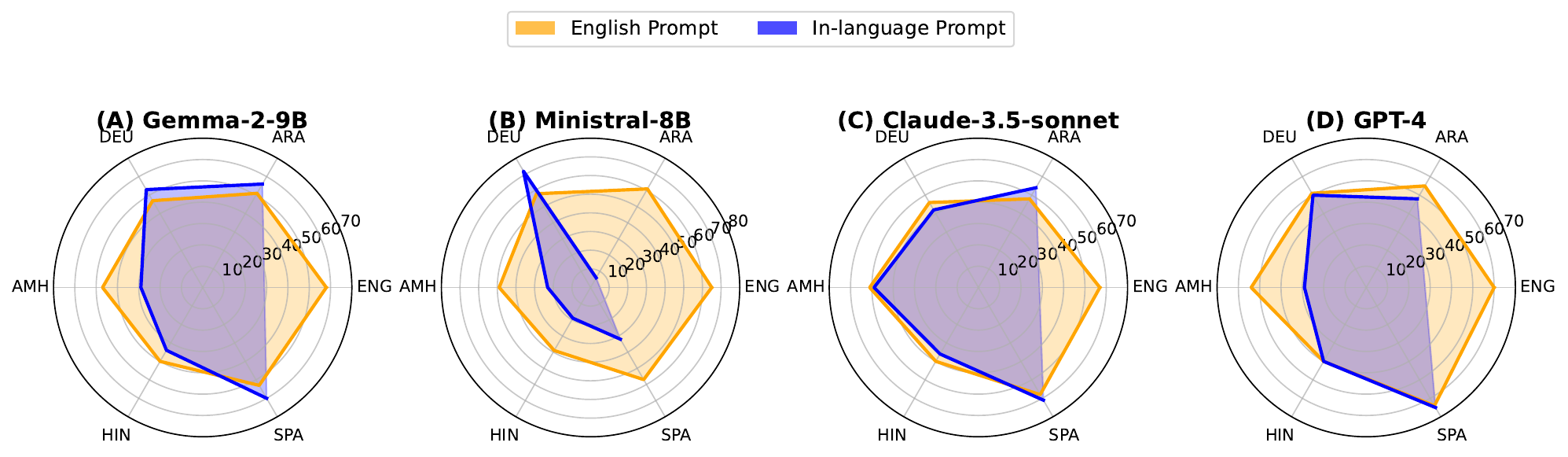}
    \caption{Emotion prediction accuracy in radar chart across countries in English and in-language prompts. For lower-resource languages, English tends to work substantially better.}
    \label{fig:summ}
\end{figure*}

\begin{table*}[]
\centering
\begin{tabular}{l|ccccccccccc}
\toprule
\multirow{2}{*}{LLMs} & USA & \multicolumn{2}{c}{UAE} & \multicolumn{2}{c}{Germany} & \multicolumn{2}{c}{Ethiopia} & \multicolumn{2}{c}{India}& \multicolumn{2}{c}{Mexico} \\
\cline{2-12}
& EN & EN & AR & EN & DE & EN & AM & EN & HI& EN & ES \\
\hline
Llama-3.2-3B&0.44 & 0.20&0.12 & 0.20&0.20 & 0.18&0.30 & 0.26&0.26 &0.28& 0.37\\
Llama-3.1-8B& 0.58 & 0.52&0.47 & 0.48&0.39 & 0.48&0.14 & 0.37&0.34 &0.56&0.57\\
\hline
Gemma-2-2B &0.62 & 0.59 &0.45&0.64 & 0.54&0.48 & 0.17&0.38 & 0.29&0.58 & 0.56\\
Gemma-2-9B &0.57 & 0.51 &\underline{0.56}&0.47 & 0.53&0.47 & 0.29&0.40 & 0.34&0.53 & 0.60\\
\hline
Aya-expanse-8b* &0.38& 0.28&0.20&0.30 & 0.24&0.30 & 0.29&0.32 & 0.29&0.37& 0.47\\
Aya-101-13B &0.60 & 0.60 & 0.43&0.68 & 0.43&0.52 & 0.34&0.39 & 0.34&0.56 & 0.48\\
Ministral-8B &\textbf{0.65} & \textbf{0.61} &0.06&0.58 & \textbf{\underline{0.72}}&0.49 & 0.23&0.39 & 0.19&0.57 & 0.32\\
\hline
Claude-3.5-sonnet &0.57 &0.48&0.54 &0.46&0.42 &0.51&0.49 &0.40&0.36&0.58&0.61\\
Claude-3-opus &0.54 &0.48&0.47 &0.43&0.32 &0.53&0.43 &0.37&0.36 &0.61&0.61\\
Gemini1.5-flash &0.56&0.56&\underline{0.56}&0.46&0.48&0.51&\underline{0.51}&\textbf{0.41}&\textbf{\underline{0.41}}&0.62&0.64\\
GPT-4 &0.60&0.55&0.48&0.51&0.50&\textbf{0.54}&0.29&0.40&0.40&0.64&\textbf{\underline{0.65}}\\
\bottomrule
\end{tabular}
\caption{LLMs’ accuracy for the \textbf{emotion prediction} task. Columns labeled EN/AR/DE/AM/HI/ES show the prompt language for each corresponding culture (e.g., UAE is tested with English and Arabic). The highest-scoring model across English and in-language prompts is highlighted in \textbf{bold}, and the best model for the in-language prompt is \underline{underlined}. * indicates a non-instruction fine-tuned model.}
\label{tab:emo}
\end{table*}

\begin{table*}[]
    \centering
    \small
    \resizebox{\textwidth}{!}{
    \begin{tabular}{p{8cm}llllll}
    \toprule
    \textbf{Questions Examples from CuLEmo}     &\textbf{US}  &\textbf{AE}  &\textbf{DE}  &\textbf{ET} & \textbf{IN} & \textbf{MX}\\
    \hline
        \multirow{2}{8cm}{How would you feel when you did not tip the waiter at the restaurant?} 
    & guilt & neutral  & neutral & guilt & guilt  & guilt \\\cline{2-7}
    &  guilt & guilt  & guilt & guilt & guilt  & guilt \\
    \hline
\multirow{2}{8cm}{How would you feel when someone insults someone’s religion?} 
    & neutral & anger & neutral & anger & neutral & anger \\\cline{2-7}
    &  anger &  anger  &  anger &  anger &  anger  &  anger \\
    \hline
    \multirow{2}{8cm}{How would you feel if someone wears black to a wedding?} 
    & sadness & neutral & neutral & anger & neutral & neutral \\\cline{2-7}
    &  neutral &  neutral   &  neutral &  neutral &  neutral  &  neutral \\
    \hline
    \multirow{2}{8cm}{How would you feel when you see a female wearing small pants on the street?} 
    & neutral & sadness & neutral & anger & neutral & neutral \\\cline{2-7}
    &  neutral &  neutral &  neutral &  neutral &  neutral  &  neutral \\
    \hline
    \multirow{2}{8cm}{How would you feel when your attendee joined the meeting after 10 minutes started?} 
    & anger & neutral & anger & neutral & anger & anger \\\cline{2-7}
    &  neutral &  anger  &  anger &  anger &  anger  &  anger \\
    \hline
    \multirow{2}{8cm}{How would you feel if your parents arranged a marriage for you without your input?} 
    & anger & anger & neutral & neutral & neutral & neutral \\\cline{2-7}
    &  anger &  anger  &  anger &  anger &  anger  &  anger \\
    \hline
    \multirow{2}{8cm}{How would you feel upon receiving the message that you have been accepted as a medical student?} 
    & joy & joy & neutral & joy & joy & joy \\\cline{2-7}
    &  joy &  joy  &  joy &  joy &  joy  &  joy \\
    \bottomrule
    \end{tabular}
    }
    \caption{Examples of emotion predictions from GPT-4 model. The first row under the country-code columns in each section represents the gold-label emotions, while the second row displays the predicted label. The predictions are using country context, by adding \texttt{You live in <<country name>>, How do you feel when ...} to the prompt.}
    \label{tab:pred}
\end{table*}

\begin{table*}[]
\centering
\resizebox{\linewidth}{!}{
\begin{tabular}{lcccccc|ccccc}
\toprule
\multicolumn{7}{c|}{\textbf{English prompt}} & \multicolumn{5}{c}{\textbf{In-language prompt}} \\
\cmidrule{2-7}\cmidrule{8-12}
 \textbf{Lang. }  & US & AE & DE & ET &IN& MX & AE & DE & ET &IN& MX \\
\midrule
Llama-3.2-3B &0.44&0.41&0.36&0.35&0.31&0.46&
0.13&0.19&0.30&0.25&0.37\\
Llama-3.1-8B &0.62&0.54&0.50&0.51&0.36&0.60&
0.46&0.31&0.18&0.32&0.56\\
\midrule
Gemma-2-2B &0.61&0.59&0.68&0.49&0.32&0.57&
0.47&0.56&0.14&0.27&0.52\\
Gemma-2-9B &0.59&0.54&0.54&0.50&\textbf{0.41}&0.57&
0.59&0.59&0.28&0.34&0.61\\
\midrule
Aya-101-13B* &0.61&0.57&0.60&\textbf{0.55}&0.38&0.58& 
0.34&0.39&0.31&0.30&0.49\\
Ministral-8B &\textbf{0.66}&\textbf{0.60}&0.56&0.50&0.38&0.59&
0.07&\textbf{0.67}&0.20&0.23&0.34\\
\midrule
Claude-3.5-s. &0.56&0.51&0.39&0.49&0.36&0.60&
0.51&0.35&0.53&0.38&0.61\\
Claude-3-op &0.52&0.40&0.37&0.49&0.36&0.59&
0.45&0.30&0.47&0.37&0.58\\
Gemini1.5 &0.53&0.51&0.39&0.48&0.39&0.59&
0.54&0.42&0.52&0.37&\textbf{0.65}\\
GPT-4 &0.58&0.55&0.46&0.48&0.37&0.63&
0.46&0.45&0.27&\textbf{0.41}&0.64\\

\bottomrule
\end{tabular}
}
\caption{Emotion predicton results using \textbf{English prompts} and \textbf{in-language prompts} without specifying the country context, \textit{"You live in «country name»"}. The language column names are the two-letter targeted countries. The \textbf{boldface} result indicates the better results for each country. * indicates a non-instruction fine-tuned model.}
\label{tab:bias}
\end{table*}

\subsection{Culture-Aware Emotion Prediction}

\begin{figure*}[]
    \centering
    \includegraphics[width=\linewidth]{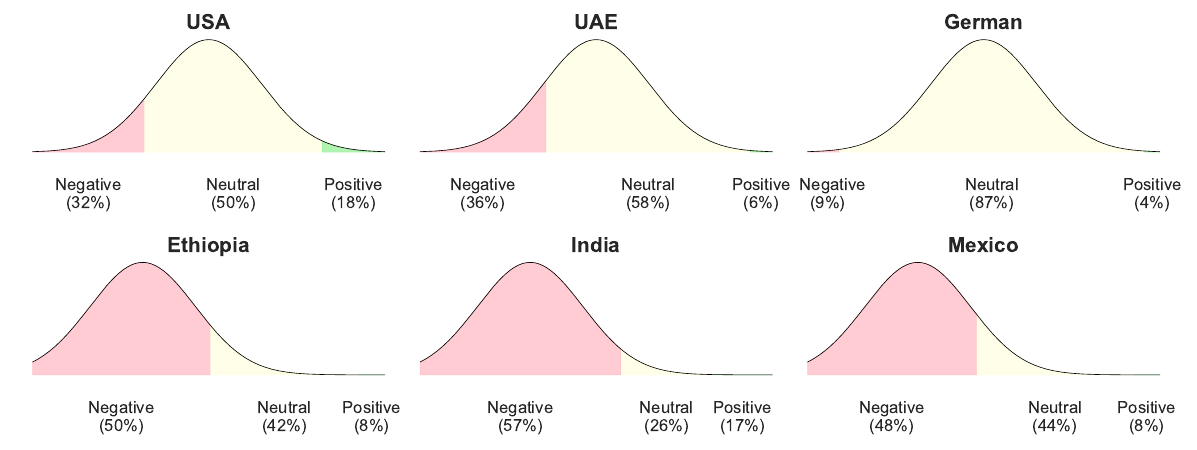}
    \caption{Sentiment (positive, negative, and neutral) distribution across countries in the CuLEmo dataset.}
    \label{fig:bell}
\end{figure*}

\begin{table*}[]
\centering
\begin{tabular}{l|ccccccccccc}
\toprule
\multirow{2}{*}{LLMs} & USA & \multicolumn{2}{c}{UAE} & \multicolumn{2}{c}{Germany} & \multicolumn{2}{c}{Ethiopia} & \multicolumn{2}{c}{India}& \multicolumn{2}{c}{Mexico} \\
\cline{2-12}
& EN & EN & AR & EN & DE & EN & AM & EN & HI& EN & ES \\
\hline
Llama-3.2-3B &0.57&0.41&0.15&0.20&0.24&0.48&0.51&0.60&0.56&0.55&0.60\\
Llama-3.1-8B &0.65&0.58&0.50&0.49&0.43&0.59&0.43&0.57&0.55&0.64&0.68\\
\hline
Gemma-2-2B &0.68&0.65&0.58&0.66&0.59&0.56&0.46&0.50&0.59&0.66&0.66\\
Gemma-2-9B &0.55&0.57&\underline{0.63}&0.49&0.56&\textbf{0.62}&0.50&0.57&0.61&0.63&0.68\\
\hline
Aya-expanse-8b* &0.45&0.33&0.26&0.31&0.27&0.39&0.50&0.45&0.58&0.45&0.57\\
Aya-101-13B &0.65&0.64&0.55&0.69&0.45&0.55&0.53&0.52&0.56&0.61&0.63\\
Ministral-8B &0.70&\textbf{0.66}&0.06 &0.60&\textbf{\underline{0.73}}&0.59&0.46&0.54&0.24&0.65&0.53\\
\hline
Claude-3.5-sonnet &0.65&0.58&0.61&0.49&0.46&0.66&\underline{0.62}&0.60&0.61&0.66&0.72\\
Claude-3-opus &0.63&0.60&0.56&0.46&0.35&\textbf{0.67}&0.58&0.59&0.61&0.72&\textbf{\underline{0.75}}\\
Gemini1.5-flash &0.63&0.63&\underline{0.63}&0.48&0.50&0.63&0.60&0.60&0.53&0.69&0.71\\
GPT-4 &\textbf{0.67}&0.61&0.54&0.48&0.51&0.62&0.49&0.58&\textbf{\underline{0.62}}&0.72&\textbf{\underline{0.75}}\\

\bottomrule
\end{tabular}
\caption{Accuracy for \textbf{culture-aware sentiment analysis} (positive/negative/neutral) with English and in-language prompts. The highest-scoring model across English and in-language prompts is highlighted in \textbf{bold}, and the best model for the in-language prompt is \underline{underlined}. * is not an instruction fine-tuned model.}
\label{tab:sent}
\end{table*}

\textbf{Do LLMs provide culturally aware emotional responses?}
A culture-aware model should accurately answer questions related to any culture, demonstrating uniformly high accuracy. To test this, we assess each LLM's emotional understanding using \textbf{English} and \textbf{in-language} prompts with the context \texttt{"You live in <<country name>>,"}. 

\noindent \textbf{Results:} Table \ref{tab:emo} presents the accuracy of each LLM for culture-aware emotion prediction. The choice of prompt language significantly influences performance. Generally, proprietary models are less affected by in-language prompts compared to open-source models, especially for Spanish and German. Certain cultures appear better represented in the models—Ministral-8B scores highly on German (72\%), and GPT-4 performs best on Mexican (65\%). In contrast, performance in Indian culture (Hindi) lags, particularly those using Hindi, Amharic, or Arabic scripts. Larger models do \textbf{not} always outperform smaller ones; Gemma-2-2B and Ministral-8B show competitive or superior accuracy relative to some proprietary models. When prompted in English, all models achieve a reasonable accuracy. Ministral-8B can exceed proprietary performance in English and German. GPT-4 answers the same emotion while using the corresponding country name in the prompting; see predicted examples in Table \ref{tab:pred}. Overall, results suggest that \textbf{culture-aware emotion understanding} remains challenging for the tested LLMs, especially for low-resource languages and cultures. 

\subsection{Culture Representation in LLMs}
\textbf{Which culture is more represented in LLMs?} Here, we evaluate LLM performance using English prompts without any explicit country context. We then measure how models respond to events from each target country. 


\noindent \textbf{Results:} According to Table \ref{tab:bias}, \textbf{English prompt} column category, the USA, Mexico, and Germany consistently achieve higher accuracy scores, while the UAE, Ethiopia, and India remain less accurately represented. This suggests that certain cultures may be more prevalent in the underlying training data.



\subsection{Does Language Represent Country?}
\textbf{Can LLMs identify one country's culture based solely on the prompt language?} In this experiment, we remove explicit country context (\texttt{"You live in <<country name>>"}) and test whether LLMs can infer cultural cues only from the language used in the prompt. 



\noindent \textbf{Results:} Table \ref{tab:bias}, \textbf{in-language prompt} column category, shows that accuracy drops significantly when country context is omitted. Comparing these scores with Table \ref{tab:emo} (where country context is included), we see consistent performance boosts (e.g. +1\% in Spanish with GPT-4, +5\% in Arabic with Gemini1.5,  +6\% in German with Ministral, +9\% in Amharic with  Claude-3.5-sonnet, and +21\% in Hindi with GPT-4) when the prompt explicitly names the country. This indicates that language alone does not reliably convey cultural context. Notably, models like Claude-3.5-sonnet and GPT-4 show improvements in Indian and Ethiopian data when country context is specified, while English prompts (without “USA” context) are less affected. Overall, providing  the country name remains crucial for accurate culture-aware emotion understanding, especially for less-resourced languages.

\subsection{Culture-Aware Sentiment Analysis}
For the sentiment analysis experiment, we follow \citet{de2019towards} and \citet{davani-etal-2022-dealing} by grouping emotions into positive (\textit{joy}), negative (\textit{fear, anger, guilt, sadness}), and neutral sentiments. Table~\ref{tab:sent} shows the accuracy of each LLM under this three-class setup. 

\noindent \textbf{Results:} Table \ref{tab:sent} shows better overall performance on \textbf{sentiment analysis} compared to \textbf{fine-grained emotion} classification. For example, GPT-4 gains notable accuracy (e.g., +22\% for Hindi). Similarly, the highest score appears for Mexican culture (Spanish) with Claude-3-opus and GPT-4, each at 75\%. By contrast, Aya-expanse-8b struggles more, as it is not instruction fine tuned. Smaller models like Gemma-2-2B are competitive with proprietary models. Still, performance drops persist for Amharic and Hindi, reflecting the challenges of culture-aware tasks in lower-resource contexts.


\section{Discussion}
\label{sec:discussion}
Our analyses provide several insights into the current state of LLMs with respect to cultural emotion understanding. We highlight three main lessons learned and propose potential steps to enhance the cultural awareness of LLMs.

\subsection{Variance Across Languages and Cultures} Data analysis from the CuLEmo dataset in Figure \ref{fig:distr} shows notable differences in how annotators from different countries perceive the same event differently. For instance, the events annotated from Germany have the highest proportion of \textit{neutral} (no emotion) labels. Figure \ref{fig:bell} further illustrates the distribution of positive, negative, and neutral sentiments across 400 instances in each country. 

\noindent \textbf{How are emotions distributed across languages and cultures?} Based on the dataset analysis of emotion distribution across languages, shown in Figure \ref{fig:distr}: German (87\%), Arabic (58\%), and English (50.5\%) data have the most \textit{neutral} (no emotion). Amharic (29.5\%), Arabic (22\%), and Spanish (21.5\%) languages have the most \textit{anger} emotion. These findings confirm that a single event can evoke distinct emotional reactions depending on the cultural background and language.

\subsection{Prompt Language Strongly Affects Cultural Emotion Understanding} 

As illustrated in the Table \ref{tab:bias} and summarized results in Figure \ref{fig:summ}, \textbf{prompt language} plays a major role in LLM performance for emotion prediction and sentiment analysis. For less-resourced languages like Amharic and Hindi, prompting in English consistently yields better results—sometimes by as much as a 20\% improvement. Conversely, \textbf{in-language prompts with explicit country context} tend to work best for high-resource languages such as German and Spanish. These discrepancies stem from differences in both linguistic coverage and instruction-following abilities learned during pre-training. One practical solution is to leverage English prompts while specifying the target country (e.g., \texttt{“You live in <<country name>>,”}).

\subsection{Performance Gaps Reflect Under-Represented cultures} We observe notably lower accuracy for Ethiopia and India in both emotion and sentiment tasks, suggesting that models may be less exposed to cultural practices and norms for these under-represented contexts. Ensuring greater diversity in training corpora is key to improving model performance for such cultures. Providing explicit country context can partially offset these gaps by nudging models to incorporate relevant cultural knowledge.

Overall, our findings underscore the importance of cultural context in developing and deploying LLMs. Beyond balanced data collection, researchers may explore culture-specific tuning or reinforcement learning from human feedback to further refine the abilities of models to interpret and respect cultural nuances.


\section{Conclusion}
\label{sec:conclusion}
In this paper, we evaluate a diverse set of state-of-the-art LLMs for their ability to predict culturally aware emotion prediction and sentiment analysis tasks. We investigate the influence of including explicit country references \texttt{“You live in <<country name>>”} and varying the query language. Our results indicate that LLMs tend to excel at culturally driven emotions that are well-represented in their training data and underperform for less represented cultures. Specifically, we find that 1) emotion is culture-dependent and can vary notably across languages and regions; 2) LLMs exhibit sizable performance gaps when tested on culture-specific emotions from under-represented locales; 3) providing explicit country context in prompts improves both emotion and sentiment prediction; and 4) sentiment analysis is better for the models, likely because it involved fewer class (positive, negative, neutral) than fine-grained emotion categories. 

Moving forward, we suggest training LLMs in approaches such as 1) enriching the training data with diverse cultural information from various sources like literature, news, and cultural databases and 2) implementing fine-tuning techniques that specifically train the LLM on prompts and datasets focused on different cultural contexts so that they can be culturally aware and able to generate responses that are sensitive to cultural nuances. We also encourage more extensive evaluation of multilingual models using benchmarks designed to measure cultural awareness alongside standard accuracy metrics. Future research could also explore the influence of annotator demographics---such as age, gender, education level, political stance, and religion---on culture-specific emotion annotation. Finally, we hope that releasing the \textbf{CuLEmo} dataset will foster further exploration into culturally nuanced NLP tasks and lead to more inclusive language models.

\section*{Limitations}
\label{sec:limitaitons}
\paragraph{Subjectivity of emotion} Emotional subjectivity remains a central challenge in emotion analysis tasks. Although annotating data via crowdsourcing such as Amazon Mechanical Turk (MTurk) is common in NLP dataset creation \cite{mohammad-etal-2018-semeval}, and despite applying strict qualification criteria for annotators, maintaining consistent annotations is difficult given the inherently subjective nature of emotions. 
\paragraph{Limited number of events} Our test comprises only 400 questions for each language, which is certainly not sufficient to capture the full cultural differences in emotional expression. 
\paragraph{Drawback of majority vote} We decide the final label of the annotations using majority vote, such as an emotion label greater than or equal to three votes from a total of five annotators per instance will pass as a final emotion label. As a general drawback of the majority vote, this will exclude the perspectives of minority votes. Modeling annotator-level data without applying the majority vote can address this. 
\paragraph{Limited emotion label space} Additional constraints arise from our decision to limit the emotion label space to six classes; including more emotion categories (e.g. \textit{surprise} or \textit{disgust}) during the annotation could yield more fine-grained insights \cite{niu2024rethinking}. Our dataset also covers only six languages/countries and comprises 400 events, which may restrict generalizability. 
\paragraph{Annotation bias} Emotion annotation is subjective in nature and can vary widely depending on personal background; it likely still has consistency issues, affecting the reliability of the evaluations.
\paragraph{Limited model evaluations} Regarding open-source LLMs, we opted to evaluate only small- (2B,3B) and medium-sized (8B, 9B, 13B) models due to resource constraints and for experimental reproducibility. While larger LLMs might achieve higher accuracies across target languages, they remain beyond the scope of our current setup. Finally, although 400 events allow for controlled experiments, evaluating models on more extensive and varied data would provide a clearer picture of their culture-aware performance.

\section*{Ethics Statement}
\label{sec:ethics}
We conducted this work with careful attention to ethical considerations involving data creation, annotation, and potential downstream impacts.

~\\ \textbf{1. Data Collection and Annotation}
\paragraph{Cultural Respect}The CuLEmo dataset was curated with input from native speakers and cultural representatives. We designed questions to capture diverse cultural norms and emotional responses without perpetuating stereotypes.
\paragraph{Consent and Compensation}We used Amazon Mechanical Turk (MTurk) and an in-house annotation platform for data labeling. Workers were informed of the task’s nature and compensated fairly at a rate of \$12/hour, which exceeds minimum-wage standards in the majority of the annotators’ countries of residence.
\paragraph{Privacy and Confidentiality}All scenario-based questions were artificially created or adapted from publicly available cultural information. No personally identifiable information was collected, and no real names or private details were used.

~\\ \textbf{2. Fair Representation and Potential Biases}
\paragraph{Under-Representation}While we included six languages (English, Arabic, Amharic, German, Hindi, and Spanish) to broaden cultural coverage, bias in representation is inevitable in such datasets and evaluations. Clearly, many global cultures and languages remain unrepresented in our work, including minority language speakers of the languages we studied and speakers of those languages in less dominant regions.  Additionally, cultures are complex and not subject to clean delineation.  We therefore make no contention that this is a complete or fully representative dataset.
\paragraph{Subjectivity of Emotions}Emotions are inherently subjective and influenced by personal and cultural backgrounds. Crowd-sourced annotations may inadvertently amplify majority cultural norms or obscure minority perspectives. We minimized these risks by providing clear guidelines, but acknowledge that subjective variation is inevitable.

~\\ \textbf{3. Responsible Use of the Dataset and Models}
\paragraph{Cultural Sensitivity}The dataset includes prompts and scenarios potentially sensitive to specific cultural contexts (e.g., religious practices, social norms). We urge researchers and practitioners to exercise cultural sensitivity and caution when using the dataset or resulting models in applications that could impact cultural or ethnic groups.
\paragraph{Downstream Applications}Models trained or evaluated on CuLEmo could be applied in contexts such as mental health or social support, potentially affecting vulnerable populations. We encourage developers to consider safety, fairness and informed consent when deploying such systems. We caution against deployment in high-stakes settings, particularly without appropriate safeguards, user testing, and especially ethical oversight.

~\\ \textbf{4. Transparency and Future Work}
\paragraph{Open Access}We release the CuLEmo dataset publicly to facilitate reproducibility and encourage further research in culturally aware NLP.
\paragraph{Ongoing Improvement}Future efforts should expand cultural and linguistic diversity, refine annotation protocols, and include more nuanced emotional labels. We welcome community feedback to improve both the dataset and modeling approaches.

We aim to advance culturally aware NLP through responsible data practices, fair representation, and transparent sharing, and hope this work fosters a more inclusive understanding of emotion across languages and cultures.

\section*{Acknowledgments}
This work is partially funded by the NAACL regional fund 2024. 
We gratefully acknowledge computing resource support provided by the Computing Research Center (CIC) at the National Polytechnic Institute (IPN) and the Marian E. Koshland Integrated Natural Sciences Center at Haverford College. We extend our sincere thanks to the Digital Scholarship team at Haverford College, particularly Anna Lacy and Patricia Guardiola, for their assistance throughout this project. We thank Professor Xerxes Minocher at Haverford College for their valuable inputs. Grissom is funded by NSF grant 2403439.

\bibliography{custom}
\onecolumn

\appendix
\section{Dataset Examples}\label{app:example}
The section below shows English examples from CuLEmo datasets.
\label{sec:appendix}
\begin{tcolorbox}[colback=black!5!white,colframe=black!75!black, title= English Examples from the CuLEmo dataset]
\begin{enumerate}[noitemsep]
\item How would you feel when you did not tip the waiter at the restaurant?
\item How would you feel when someone insults someone's religion? 
\item How would you feel if someone wear black to a wedding?
\item How would you feel when you see a female wearing small pants on the street
\item How would you feel when your attendee joined the meeting after 10 minutes started
\item How would you feel when you saw a young dog running to you?
\item How would you feel when someone accepted your gift in one hand?
\item How would you feel when you crossed an older man in the street?
\item How would you feel when someone calls a religious leader by their name
\item You are a doctor. How would you feel when someone calls you by your name without the doctor prefix
\item How would you feel if your parents arranged a marriage for you without your input?
\item How would you feel if someone touched or shook hands with you while you were of the opposite sex?
\item How would you feel when a guest leaves food on their plate at your home? 
\item How would you feel if you saw a woman not cover her hair while entering a religious place?
\item How would you feel when you were expected to wash other legs
\item How would you feel if someone refused your gift before accepting it?
\item How would you feel when you kissed your girlfriend/boyfriend and saw your father?
\item How would you feel when your parent caught you with a pornographic film?
\item If someone is passing near to you, how would you feel when they didn't give you a greeting?
\item How would you feel when you see a female eating on the street?
\item How would you feel when you see a Sheikh drinking alcohol?
\item How would you feel when you see a priest drinking alcohol?
\item How would you feel when you see homosexual friends?
\item How would you feel when someone shows you his middle finger gesture?
\item How would you feel when someone gave the money in their left hand?
\item How would you feel if your guests chose to keep their shoes on when entering your home?
\item How would you feel if a friend offered to pay for your meal in a restaurant?
\item How would you feel if your child expressed a desire to marry someone from a different religion?
\item How would you feel if someone bargained for your prices?
\item When you are at a meal, how would you feel about being told to wait for the host to say start before starting to eat?
\item How would you feel when you expected to wait for the eldest person to start eating before you begin?
\item How would you feel if someone were scrolling their phone during a meal?
\item How would you feel when you saw a young dog running toward you?
\item How would you feel when someone cheek-to-cheek kisses your wife/husband?
\item How would you feel if someone bargained for your prices?
\item How do you feel when a guest arrives late to your lunch invitation?
\item How do you feel seeing someone eating in hand without using utensils?
\item How would you feel if someone called an elder by their first name without a title?
\item How would you feel if a guest left your home without eating anything?
\item How would you feel if your friend didn't stand up when an elder entered the room?
\end{enumerate}
\end{tcolorbox}

\section{Implementation Details}
\subsection{Hyper-parameters}
For open-source LLMs, we used the default generation hyperparameters (top-p sampling with p = 0.9 and temperature = 0.0, max\_new\_tokens=200). For others, we directly employed their pre-defined interfaces, either through their online API or the CHAT function from the Transformers library. The proprietary models are called through their API in Python. All open-source models are evaluated using default parameters from the hugging face.

\subsection{LLM Versions}
Versions of the proprietary  LLMs and hugging face names for the open-source LLMs are given below.
\begin{itemize}
    \item Llama-3.2-3B \cite{llama3.2} — meta-llama/Llama-3.2-3B-Instruct
    \item Llama-3.1-8B \cite{llama3.1} — meta-llama/Llama-3.1-8B-Instruct
    \item Gemma-2-2B  \cite{gemmateam2024gemma} — google/gemma-2-2b-it
    \item Gemma-2-9B  \cite{gemmateam2024gemma} — google/gemma-2-9b
    \item Aya-expanse-8b \cite{dang2024ayaexpanse} — CohereForAI/aya-expanse-8b
    \item Aya-101 \cite{aya-2024} — CohereForAI/aya-101
    \item Ministral-8B \cite{ministral} — mistralai/Ministral-8B-Instruct-2410
    \item Claude-3.5-sonnet \cite{claude3} — claude-3-5-sonnet-20240620
    \item Claude-3-opus \cite{claude3} — claude-3-opus-20240229
    \item Gemini1.5-flash \cite{gemini1.5} — gemini-1.5-flash-002
    \item GPT-4 \cite{gpt4technicalreport} — gpt-4 (turbo-2024-04-09)
\end{itemize}

\section{Prompts}\label{app:prompts}
\begin{figure}[hbt!]
    \centering
    \includegraphics[width=1\linewidth]{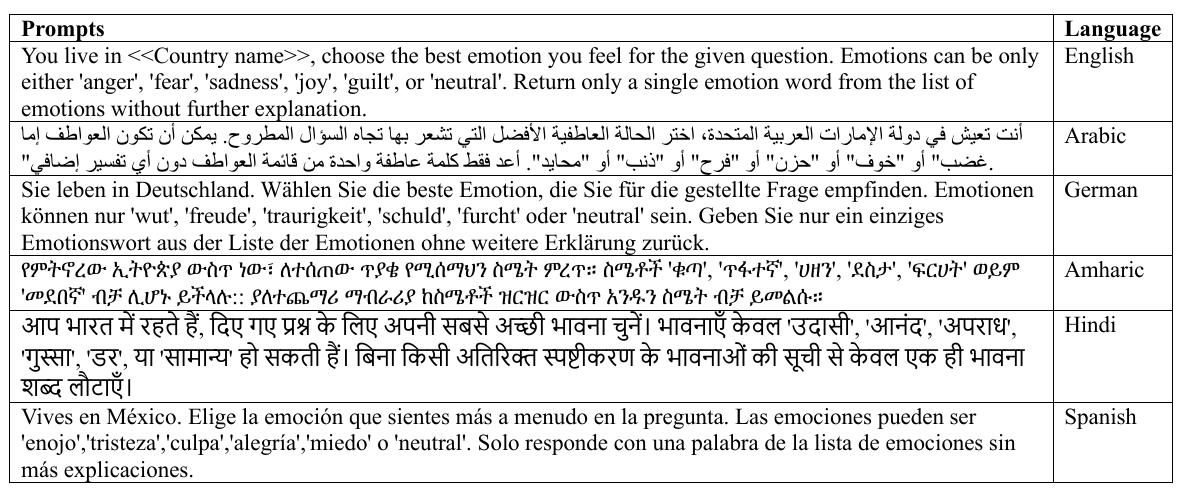}
    \captionof{table}[]{Prompts used for probing emotions from LLMs. In English prompt, <<Country name>> will change accordingly from lists of countries [United States of America (USA), Ethiopia, United Arab Emirates (UAE), Germany, India, Mexico] based on the culture prompting.  We enforce the model in the prompt to answer only one of the given options.}
    \label{fig:prompt}
\end{figure}

\clearpage
\section{English Prompt Results}
\begin{figure*}[!h]
    \centering
    \includegraphics[width=\linewidth]{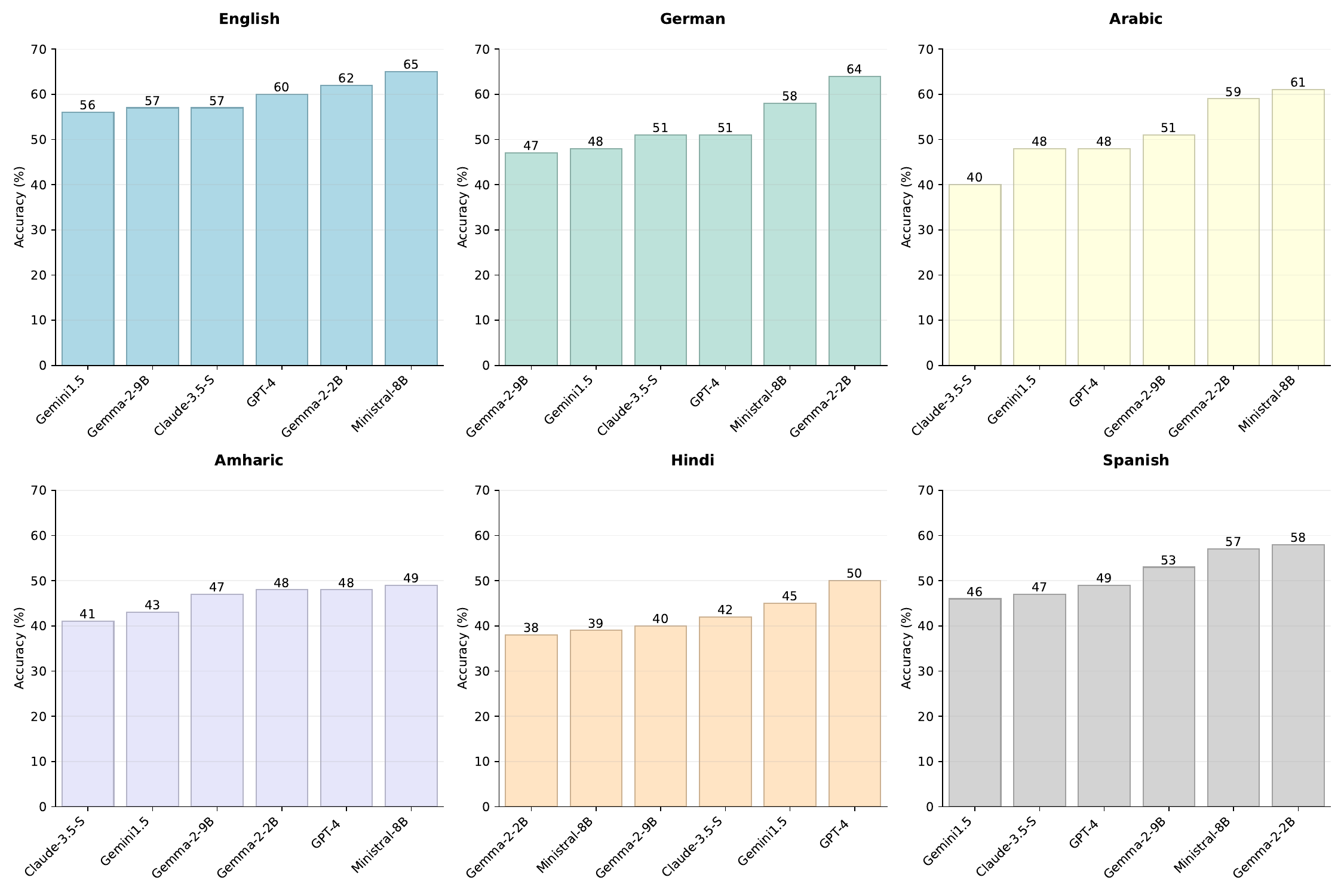}
    \caption{English prompting results with \textit{You live in <<country name>>} context for the across languages and LLMs.} 
    \label{fig:eng-prompt}
\end{figure*}

\section{MTurk Annotation Qualification Settings}
To target suitable workers on MTurk, we set the following qualifications:
\begin{enumerate}[nosep]
    \item \textbf{Location} must be in the target country for each language by assuming annotators that live in the specified country are native or adopted the culture.
    \item \textbf{Number of HITs approved} must exceed 1,000 to ensure experienced workers.
    \item \textbf{HIT approval rate} must be at least 99\%, favoring high-quality, consistent annotators
\end{enumerate}

\end{document}